# Textural Approach for Mass Abnormality Segmentation in Mammographic Images

Khamsa Djaroudib[1], Abdelmalik Taleb Ahmed[2] and Abdelmadjid Zidani[3]

[1] LaStic Laboratory, Faculty of Science, Computer Science Department, Batna University, (05000) Algeria.

[2] LAMIH Laboratory, Valenciennes University, UVHC, (59300) France.

[3] LaStic Laboratory, Faculty of Science, Computer Science Department, Batna University, (05000) Algeria.

**Abstract**

Mass abnormality segmentation is a vital step for the medical diagnostic process and is attracting more and more the interest of many research groups.
Currently, most of the works achieved in this area have used the Gray Level Co-occurrence Matrix (GLCM) as texture features with a region-based approach. These features come in previous phase for segmentation stage or are using as inputs to classification stage.
The work discussed in this paper attempts to experiment the GLCM method under a contour-based approach. Besides, we experiment the proposed approach on various tissues densities to bring more significant results. At this end, we explored some challenging breast images from BIRADS medical Data Base. Our first experimentations showed promising results with regard to the edges mass segmentation methods.
This paper discusses first the main works achieved in this area. Sections 2 and 3 present materials and our methodology. The main results are showed and evaluated before concluding our paper.

## 1. Introduction

Diagnosis and cure of breast cancer depend strongly of on early detection and treatment of abnormalities (mass or micro-calcification) of breast. And breast cancer is the first cause of death by cancer at the women. Unfortunately, due to the large variability of size, shape and margin, and its confusion with the mammary tissue, mass abnormality detection and/or segmentation is still a very difficult task for the researchers more than micro-calcification abnormalities detection. Computer Aided Diagnosis (CAD) being an effective tool for radiologist [1][2][3], for giving a second and more reliable opinion in diagnosis. The detection and/or segmentation is the first and key stage in the complete process of CAD [4][5].

Masses are characterized by their location, size, shape and margin [6][7] and the large variation in size and shape in which masse can appear, make mass segmentation a challenging task for researchers. In additional, at the most of cases, mammograms exhibit poor image contrast tissue density (fatty, dense or glandular), then tissue can overlap with breast tumor region [8] as the mass abnormality [9]. According to these problems, many mass segmentation and/or detection methods are developed. We can see review and recent advance of them in [10] [4], [11], [12]. For example, pixel based methods [13] [14] [15], such as region growing and its extensions; region based methods [16][17], e.g., filter based methods; and simple edges based methods [18], e.g., the gradient filters, are employed widely in the early stage for mass segmentation. Though these types of methods are easily to implement, it is still difficult to acquire satisfied segmentation results for masses of ambiguous boundaries. This is because simple feature cannot handle the complex density distributions and topologies of the masses and normal breast tissue. To find more accurate boundaries of masses, some researchers use active contour methods [19][20][21], the efficiency of depends for adjusting parameters.

Many methods cited above use texture information, because textures features are more rich information in segmentation process [22][23] and specially in medical images [24], these have been proven to be useful in differentiating mass and normal breasts tissues. The authors in [26] show that the area of a tumor exhibit typically low texture compared to normal parenchyma, and the authors in [8] concluded that the texture features demonstrate more prominent differences between tumor and normal tissues than the intensity feature. In this idea, most methods include textures features use GLCM in segmentation or classification stage of CAD [27] and most of them use GLCM in region approaches, in order to extract texture features in previous stage for mass segmentation or classification stages, we cite some examples in [28][29][30][31]. However, there is no significant work which used these matrices in an edges approach. In this paper, we contribute and propose a mass segmentation method by edges detection approach, based on GLCM in order to extract textures images representing textures parameters. Our idea is based on the fact that variance or contrast parameter can detect the spatial change between mass and non mass tissue in region border. Then texture descriptor as the contrast extract from GLCM is compute in each pixel in ROI (Region of Interest) image give an important information to detect edges mass contours.

Our approach split in two stages. At first, we applied smoothing (denoising) and enhancing method to enhance breast image [32]. Respectively, an anisotropic filter diffusion SRAD (Speckle Reducing Anisotropic Diffusion) [33] and Contrast-limited Adaptive Histogram Equalization (CLAHE) are used. Second, for each pixel in a ROI, a contrast descriptor is computed from the co-occurrence matrix of the pixels, and the contrast image is obtained. Mass contour is identified. We applied the proposed algorithm to some challenging breast images in BIRADS database including poor contrast tissue density (fatty, dense or granular) and the segmented mass done by our algorithm is compared to segmentation carried by an expert radiologist by measuring Dice coefficient, F-measure and area under the curve (Az).

## 2. Materials and Data description

Our method was applied on the mini-MIAS dataset (http://peipa.essex.ac.uk/info/mias.html). It is available online freely for scientific purposes. The films were digitized and the corresponding images were annotated according to their breast density by expert radiologists, using three distinct classes: Fatty (F), Fatty-Glandular (G) and Dense-Glandular (D). Any abnormalities were also detected and described, including calcifications, well-defined, spiculated or ill-defined masses, architectural distortion or asymmetry. Each pair of images in the database is annotated as Symmetric or Asymmetric. The severity of each abnormality is also provided, i.e., benignancy or malignancy.

## 3. Methodology: GLCM for edges mass detection and segmentation

The main steps of proposed methodology are summarized in Fig. 1:

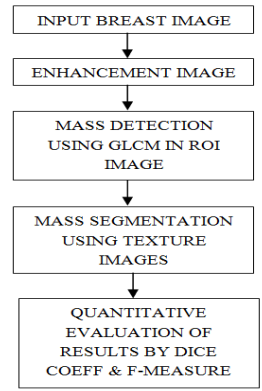

Fig. 1 Steps for the mass segmentation methodology

### 3.1 Enhancement Images

The performance of methods based on texture information is highly dependent on the pre-processing (enhancement) of the input image [26], so many researchers focus in this stage of CAD.

For our approach, this stage is our key to have the best results for the mass segmentation stage. Most mammogram images have low intensity contrast, then we applied smoothing (denoising) and enhancing method to enhance breast image [32]. We suggested applying respectively, the anisotropic filter diffusion SRAD (Speckle Reducing Anisotropic Diffusion) [33] and Contrast-limited Adaptive Histogram Equalization (CLAHE) for enhancing image.

Instead of most studies, in our approach and in the aim to perform texture information, denoising and enhancing steps are applied in whole breast image and then, we extract suspicious ROI image. So, our SRAD algorithm can take speckle for every image independently of another one which makes this approach is more efficiency for image speckle reducing.

Images in Fig.2, show an example for input image and enhancing image with delimited ROI, and then zoom of ROI extraction image.

We used the YU scripts for SRAD [33] and results of this step are shown at Fig.2. In this figure, the image of enhancement show clearly more regions in the breast image. The clear regions are even clearer, which can correspond to a region of the masses tissue, and the dark regions are darker, what can correspond to the regions of the normal tissue (without mass).

Besides in Fig.2, we showed an image mdb004 which represents the most difficult case for the detection of the mass in the clear normal tissue, which is the dense tissue.

For other cases, the images are even more contrasted to improve the next stage of our methodology which is the computing of the images of texture.

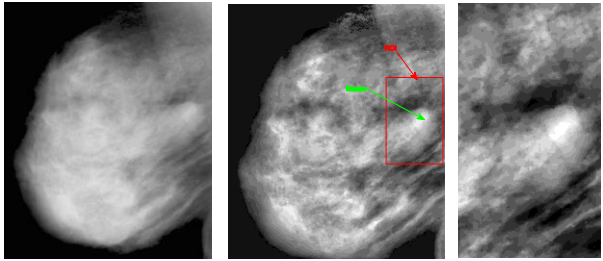

Fig. 2 Breast image mdb004 input (in left), mdb004 enhancement (center) show ROI (red color) and mass (green color), image zoom of ROI (in right)

### 3.2 Edges mass detection by computing GLCM

In ROI image, we compute GLCM according to three important parameters: direction (angle), a neighbourhood size and texture descriptor of Harralick [34].

#### 3.2.1 *Compute direction:*

Compute one angle 0°, 45°, 90°, 135° do not give closed outlines, then we compute all directions and calculate their sum, see Fig.3.

Fig. 3, is an example of Brodatz image. We show images of texture which is contrast descriptor of Harralick [34], in direction 0°, 45°, 90°, 135° and image representing the sum of these four images. In the image sum, we see clearly more closed outlines.

Fig.4, is our mammographic ROI image of breast mdb004. We confirm so the remark on the Brodatz image of Fig.3.

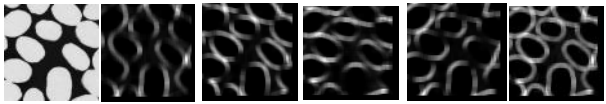

Fig. 3 From left to right : Brodatz D75 image, Image Contrast in 0°, Image Contrast in 45°, Image Contrast in 90°, Image Contrast in 135° and Image Contrast Sum (0°+ 45°+ 90°+135°).

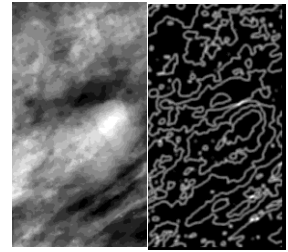

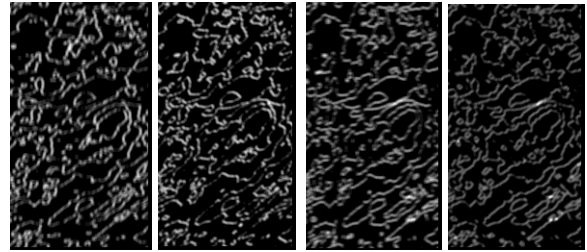

Fig. 4 Top: in left, mdb004 image input, in right, the image sum of four directions contrast image which shows closely contours.
Bottom: Four images of contrast , from left to right, respectively in 0°, 45°, 90°, 135°

#### 3.2.2 *Compute a neighbourhood size:*

For synthetic Brodatz images, we can see on Fig.5 that in mask 3x3, edges are more smooth than mask 7x7 and 9x9. The detected edges are more fuzzy if the neighbourhood size is big. But in reality, the choice of the neighbourhood size depends on textures of objects in image. For the images of mammography, neighbourhood in mask size of 7x7 give better smooth edges than mask size of 3x3 and finer outlines than mask size of 9x9.

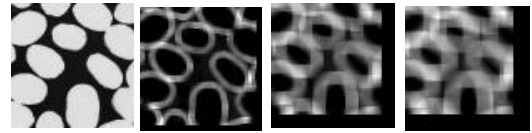

Fig. 5 From left to right: Brodatz D75 image, Image Contrast in mask 3x3, Image Contrast in mask 7x7 and Image Contrast in mask 9x9.

#### 3.2.3 *Compute texture descriptor:*

Instead of taking the most known four descriptors extracted from GLCM (1.entropy, 2.contrast, 3.second angular moment and 4.inverse differential moment), we take only the contrast descriptor, which measures the heterogeneity of an image and detect spatial variations of grey level intensity in image. Besides, it can summarize all the information of texture we needs.

To extract texture images, and according to these previous parameters (direction, neighborhood size), we compute contrast descriptor of Harralick [34]. Each pixel of ROI image is replaced by this descriptor.

In this work, we use "MatlabR2008b" formulation of contrast descriptor, *Eq. (1)*:

$$contrast = \sum_{i,j} |i-j|^2 p(i,j) \qquad (1)$$

This equation returns a measure of the intensity contrast between a pixel p and its neighbourhood in the size of 7x7. Then our algorithm computes this descriptor over the whole ROI image.

## 4. Experimental results

4.1 Edges mass detection in different densities of tissue

We applied the proposed approach to BIRADS database of breast images, on Mini-MIAS dataset. Tests are done in images with different densities of tissue, fatty, dense or glandular. For each image, contrast descriptor of Harralick is computed. We obtain the contrast texture images where the mass is identified by its borders.

In Fig.6, the examples of three images mdb004, mdb005, mdb019 which represent respectively: in first an breast image with a dense tissue, regions are of clear white color on the images of mammography; in second an image with a fatty tissue, regions are of dark grey color on the images; and finally an image with a glandular tissue, regions are of mixed color, clear white time and grey dark on the mammographic images. This information is given according to the annotations of the MIAS database.

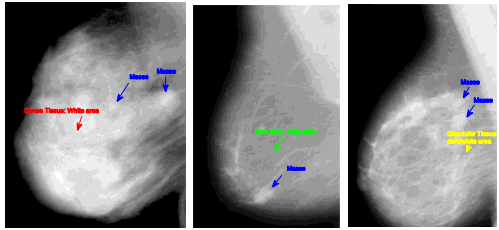

Fig. 6 From left to right: mdb004 breast, mdb005 breast, mdb019 breast. Blue color: masses, red color: dense tissue, green: fatty tissue, yellow color: Glandular tissue

4.2 Quantitative evaluation of mass edges detection

For evaluating edge detection, we select identified mass contours, according to expert image and we compute Dice Coefficient at first. Second we compute Precision, Recall and F-measure in order to calculate the area under the curve (Az). Then a comparison with an example of the recent results in the literature is given. It is results obtained for various methods of detection and/or segmentation of masses.

*4.2.1 Quantitative evaluation with Dice Coefficient:*

The segmented mass is compared to segmentation carried by an expert radiologist by measuring Dice coefficient.

Our approach was applied and tested in the challenging images of the mini-MIAS dataset. We show here the most representative and speaking cases. We quote, the cases where the tissue is dense (e.g. Fig.7), the cases where the tissue is fatty (e.g. Fig.8) and the cases where the tissue is glandular (e.g. Fig.10).

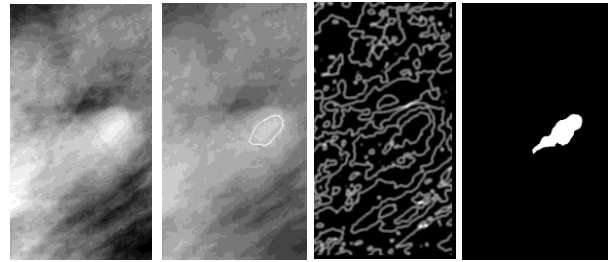

Fig. 7 In dense tissue: From left to right: ROI image, ROI image with borders (whites) of the mass region by the expert radiologist, image ROI of the texture with mass detection, image mass segmentation by applied mask, Dice Coeff = 93.39%

Fig. 7 shows the case where tissue surrounding region of mass is dense. A good Percentage of resemblance with expert radiologist segmentation is compute by Dice Coefficient 93.39%. This will certainly help the expert to interpret better the shape of the mass such as detected.

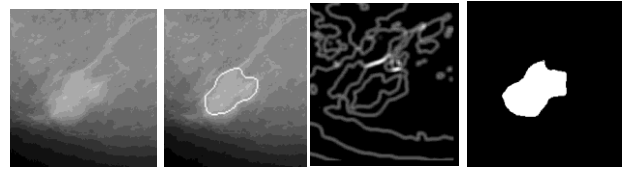

Fig. 8 In Fatty tissue: From left to right: ROI image, ROI image with borders (whites) of the mass region by the expert radiologist, image ROI of the texture with mass detection, image mass segmentation by applied mask, Dice Coeff. = 97.74%

Fig. 8 shows the case where tissue surrounding region of mass is Fatty. Here, other borders inside the region of mass delimited by the expert are detected by our method. If we follow the expert, the coefficient of resemblance will be very good 97.74%, otherwise it will not be satisfactory 69.35% in Fig. 9.

Fig. 9 shows the case where we take the mask detected inside mass region, with Dice coeff=69.35%.

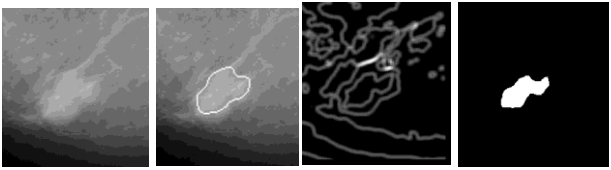

Fig. 9 From left to right: ROI image, ROI image with borders (whites) of the mass region by the expert radiologist, image ROI of the texture with mass detection, image mass segmentation by applied mask, Dice Coeff. = 69.35%.

Fig.10 and Fig.11 show the case where tissue surrounding region of mass is Glandular. The same comment as Fig 8: other borders inside the region of mass delimited by the expert are detected by our method, here the rate of resemblance with the demarcations of the expert is 86.18%, see Fig.11. But if we take the borders following the borders delimited by expert, the rate of resemblance remains good 96.67%.

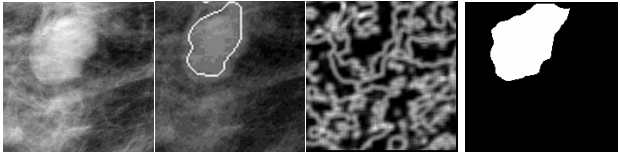

Fig. 10 In Glandular tissue: From left to right: ROI image, ROI image with borders (whites) of the mass region by the expert radiologist, image ROI of the texture with mass detection, image mass segmentation by applied mask, Dice Coeff. = 96.67%.

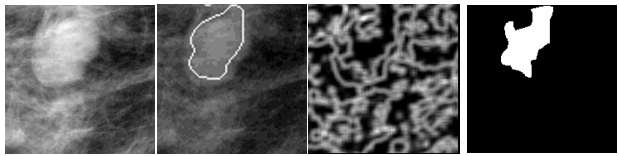

Fig. 11 From left to right: ROI image, ROI image with borders (whites) of the mass region by the expert radiologist, image ROI of the texture with mass detection, image mass segmentation by applied mask, Dice Coeff. = 86.18%.

### 4.2.2 *Quantitative evaluation with F-measure:*

We compute another evaluation in order to compare with other methods of detection and/or segmentation of mass abnormality, by area under the curve Az.

We quantify *TP, FP, TN* and *FN* as:

- *TP: True Positive*, means region segmented as mass that proved to be mass.
- *FP*: *False Positive*, means region segmented as mass that proved to be not mass.
- *FN*: *False Negative*, means region segmented as not mass that proved to be mass.
- *TN*: *True Negative*, means region segmented as not mass that proved to be not mass.
- *TPR*: *True Positive Rate*, Eq. (2)
- *FPR: False Positive Rate, Eq. (3)*

$$TPR = \frac{TP}{TP + FN} \qquad (2)$$

$$FPR = \frac{FP}{FP + TN} \qquad (3)$$

Tab. 1 shows the Precision *Eq. (4),* Recall *Eq. (5)* and F-measure *Eq. (7)* for the three cases of images cited above.

The area under the curve Az was computed with Sensitivity *Eq. (5)* and Specificity *Eq. (6).*

$$\Pr ecision = \frac{TP}{TP + FP} \qquad (4)$$

$$\operatorname{Re} call = Sensitivity = TPR \qquad (5)$$

$$Specificity = 1 - FPR \qquad (6)$$

$$F - measure = \frac{2}{(1/\Pr ecision + 1/\operatorname{Re} call)} \qquad (7)$$

| Image | Precision | Recall | F-measure |
|---|---|---|---|
| mdb004, mass in dense tissue | 0.9978 | 0.9933 | 0.9956 |
| mdb005, mass in fatty tissue | 0.9983 | 0.9781 | 0.9881 |
| mdb0019, mass in Glandular tissue | 0.9997 | 0.9375 | 0.9676 |

Tab. 1 Quantitative evaluation of the segmentation of breast mass, in training images mdb004, mdb005 and mdb019.

Then, we compared with the works of Arnau Oliver in [4], who summarized recently all the methods of mass detection and/or segmentation and who gives the obtained better results, applied also on MIAS Database. These results calculated by area under de curve Az and is between Az=0.751 and Az=0780. For our method, area under the curve Az=0.81.

## 5. Discussion and Conclusion

The stage of the detection and/or the segmentation of cancerous anomalies in the mammographic image is the key step to determine performances of CAD systems. However, the low contrast of the mammographic images and the breast tissues complexity (as well as visually and quantitatively), made that until now, most hard task is really the discrimination between the mammary tissue and the abnormality. The more the tissue is white in mammographic image (dense), the more the confusion increases.

In this study, we contributed to discriminate between normal and mass abnormality tissues by using a texture descriptor (contrast descriptor) given by the GLCM with contour-based approach, while the majority of the similar works used the texture features in a region-based approach. We also contributed to clear up the cases for three types of tissues: dense, fatty and glandular, which are most often present in the breast image.

Our results given by the Dice coefficient, F-Measure and area under the curve (Az) are good by comparing to another recent similar works and they are promising for future researches.

Detecting and/or segmenting mass edges may help radiologist experts to find size, shape and margin of mass, which are very important for decision process that leads to classify a mass as benign or malign cancer. For this, the proposed approach is especially easy and fast in terms of response time for the radiologists. The contrast descriptor allows the mass margins detecting and so her shape.

We also concluded that the enhancement stage may also be considered as a key stage in our approach. By using SRAD and CLAHE in a different way, or other enhancement methods, would give less successful and different results.

Finally, we can say that texture information is necessary for removing the ambiguity between the regions of the anomalies and the healthy regions. However, till now, the texture is hard to express under a mathematical formalism and this area of research is still open to contributions.

**Khamsa Djaroudib** received a Master in Computer Science in 1993, at the High Commissariat of Research (HCR, Algiers). She is currently a professor Assistant at the University of Batna (Algeria) and member of LaStic laboratory. Her thesis research work is mainly concerned with computer vision and medical image processing.

**Abdelmalik Taleb Ahmed** is HDR from university of Opal Côte, Calais, France. He is currently a Professor at the University of Valenciennes, France.

**Abdelmadjid Zidani** is a titular of a PhD thesis in computer science since 2002. He is currently a Professor at the University of Batna (Algeria), and member of LaStic research laboratory. His research work is mainly focused on computer human interaction (CHI) within medical settings, and especially computer supported collaborative work, medical images processing, etc.